\documentclass[letterpaper, 10 pt, conference]{ieeeconf}  

\IEEEoverridecommandlockouts                              

\usepackage{graphics} 
\usepackage{epsfig} 
\usepackage{mathptmx} 
\usepackage{times} 
\usepackage{amsmath} 
\usepackage{amssymb}  

\usepackage[utf8]{inputenc} 
\usepackage[T1]{fontenc}    
\usepackage{hyperref}       
\usepackage{url}            
\usepackage{booktabs}       
\usepackage{amsfonts}       
\usepackage{nicefrac}       
\usepackage{microtype}      
\usepackage{lipsum}
\usepackage{tikz}
\usepackage{multirow}
\usepackage{amsmath}
\usepackage{graphics}
\usepackage{subcaption}
\usepackage{makecell}
\usepackage{physics}
\usepackage{hyperref}
\usepackage{xcolor}
\usepackage{rotating}
\usepackage{pgfplots}
\pgfplotsset{compat=1.8}
\usepackage{eurosym}      
\usepackage{filecontents} 
\usepackage{tabu}
\usepackage{tikz}
\usepackage{rotating}
\usepackage{graphicx}
\usepackage{gensymb}
\usepackage{todonotes}
\usepackage{subcaption}
\usepackage{cite}
\usepackage{amssymb}
\usepackage{commath}
\usepackage{pifont}
\usepackage{float}
\newcommand{\cmark}{\ding{51}}
\newcommand{\xmark}{\ding{55}}
\graphicspath{{./figs/}}

\usepackage{acronym}
\acrodef{GTA V}[GTA V]{Grand Theft Auto V}
\acrodef{UE4}[UE4]{Unreal Engine 4}
\acrodef{CARLA}[CARLA]{Car Learning to Act}
\acrodef{PreSIL}[PreSIL]{Precise Synthetic Image and LiDAR}
\acrodef{SLAM}[SLAM]{Simoutaneous Localization and Mapping}
\acrodef{ICP}[ICP]{Iterative Closest Point}
\acrodef{VRU}[VRU]{Vulnerable Road User}
\acrodef{FPA}[FPA]{First Peak Averaging}
\acrodef{CP}[CP]{Closest Point}
\acrodef{LPIPS}[LPIPS]{Learned Perceptual Image Patch Similarity}
\acrodef{mAP}[mAP]{Mean Average Precision}
\acrodef{CD}[CD]{Chamfer Distance}
\acrodef{LPCS}[LPCS]{Learned Point Cloud Similarity}
\acrodef{FOV}[FOV]{Field of View}
\acrodef{SMPL}[SMPL]{Skinned Multi-Person Linear model}
\acrodef{IDW}[IDW]{Inverse Distance Weighted}

\title{\LARGE \bf
PCGen: Point Cloud Generator for LiDAR Simulation
}

\author{Chenqi Li, Yuan Ren and Bingbing Liu\\ 
Huawei Noah's Ark Lab, Toronto, Canada\\
\texttt{\{chenqi.li, yuan.ren3, liu.bingbing\}@huawei.com}
}

\begin{document}

\maketitle

\begin{abstract}
Data is a fundamental building block for LiDAR perception systems. Unfortunately, real-world data collection and annotation is extremely costly \& laborious. Recently, real data based LiDAR simulators have shown tremendous potential to complement real data, due to their scalability and high-fidelity compared to graphics engine based methods. Before simulation can be deployed in the real-world, two shortcomings need to be addressed. First, existing methods usually generate data which are more noisy and complete than the real point clouds, due to 3D reconstruction error and pure geometry-based raycasting method. Second, prior works on simulation for object detection focus solely on rigid objects, like cars, but \ac{VRU}s, like pedestrians, are important road participants. To tackle the first challenge, we propose \ac{FPA} raycasting and surrogate model raydrop. \ac{FPA} enables the simulation of both point cloud coordinates and sensor features, while taking into account reconstruction noise. The ray-wise surrogate raydrop model mimics the physical properties of LiDAR's laser receiver to determine whether a simulated point would be recorded by a real LiDAR. With minimal training data, the surrogate model can generalize to different geographies and scenes, closing the domain gap between raycasted and real point clouds. To tackle the simulation of deformable \ac{VRU} simulation, we employ \ac{SMPL} dataset to provide a pedestrian simulation baseline and compare the domain gap between CAD and reconstructed objects. Applying our pipeline to perform novel sensor synthesis, results show that object detection models trained by simulation data can achieve similar result as the real data trained model. 
\end{abstract}

\section{INTRODUCTION}
\label{sec:introduction}
The success of deep learning is deeply rooted in the availability of large-scale, high-fidelity datasets. Pioneering datasets \cite{geiger2013vision, caesar2020nuscenes, sun2020scalability, chang2019argoverse, wilson2021argoverse, huang2018apolloscape, houston2020one, mao2021one, choi2018kaist, patil2019h3d, xue2019blvd, maddern20171, behley2019semantickitti} facilitate development of cutting edge visual recognition systems, providing challenging benchmarks for the community. However, collection and annotation of data in the real world is very inefficient, slow and uneconomical. 
Simulation, on the other hand, give users the flexibility to generate diverse scenarios with ease, as well as providing automatically generated ground truth annotations. For LiDAR simulation, two distinct approaches have been explored: graphics engine based and real data based. Results show that real data based methods produces simulation data with lower domain gap compared to graphics engine based methods.  This paper makes the following contributions:
\begin{itemize}
  \item We present \ac{FPA} raycasting to simulate LiDAR point clouds and sensor features, accounting for noise in the reconstructed scenes. 
  \item We develop the surrogate model of a single laser head and use it for raydrop. Comparing to the UNet based raydrop method, the proposed method is scene-independent. The surrogate model of a specific LiDAR can be trained once and used in different scenes.
  \item We perform novel sensor synthesis with our simulation pipeline. The test results show that it provides high-fidelity data for the new sensor configuration, achieving similar result as the model trained on the real data.
  \item We propose \ac{LPCS} metric to measure domain gap between real and simulation point clouds, from the perspective of perception models
  \item We provide a baseline pedestrian simulation result, using \ac{SMPL} and reconstructed human models
\end{itemize}

\section{Related Work}
\subsection{Graphics Engine Based LiDAR Simulator}
Initial attempts to LiDAR simulation were spearheaded by \ac{CARLA} \cite{dosovitskiy2017carla}. Building on top of \ac{UE4} \cite{sanders2016introduction}, \ac{CARLA}'s simulation platform allows the user to customize scenarios, including agent model, density, interaction with the world, weather conditions, and sensor suite. Similarly, Yue et al. leveraged the popular, high-fidelity simulation of \ac{GTA V} to automatically extract point cloud with ground truth labels \cite{yue2018lidar}. The framework also enables users to construct diverse, customized scenarios interactively to test neural network performance in corner cases. Experiments have shown that retraining with additional synthetic point cloud significantly improves model's performance on KITTI dataset \cite{wu2018squeezeseg, wu2019squeezesegv2}. This work is further extended by the \ac{PreSIL} dataset, which improves the raycasting functionality within \ac{GTA V} to address the issues of approximating human with cylinders and missed ray-scene collisions \cite{hurl2019precise}. \ac{PreSIL} provides a large simulated dataset in KITTI format, and demonstrate that it can boost the performance on state-of-art model in KITTI 3D Object Detection benchmark.

\subsection{Real Data Based LiDAR Simulator}
Generating CAD model assets and complex scenarios are labor-intensive, making simulation difficult and costly to scale. Furthermore, domain gap between noiseless simulation and real-world data leads to poor model performance if only trained on simulation data, prompting the development of domain adaptation techniques \cite{sun2017correlation,morerio2017minimal,wu2019squeezesegv2}. Fang et al. investigated a hybrid, data-driven approach to point cloud generation framework, which combines real world scanned background point cloud and synthetic foreground objects \cite{fang2020augmented}. They show that by augmenting the real dataset with synthetic frames, instance segmentation and object detection performance is improved. Concurrently, LiDARsim employed a similar approach, leveraging real data to reconstruct both background and foreground objects \cite{Manivasagam_2020_CVPR}. LiDARsim further extended the simulation with the addition of a learning system to model the physics of LiDAR raydrop, closing the gap between real and simulation point clouds. They show that the simulation data trained models can obtain similar performance in object detection and semantic segmentation as models trained using real data, without domain adaptation techniques. Similar to LiDARsim, Langer et al. developed a simulation pipeline for domain transfer in the context of semantic segmentation \cite{langer2020domain}. Their results show that closest point raycasting, along with geodesic correlation alignment, successfully generated simulation data to adjust a model trained on the source domain (Velodyne-64) to the target domain (Velodyne-32). 
\begin{figure}[!ht]
    \centering
    \includegraphics[width=0.41\textwidth]{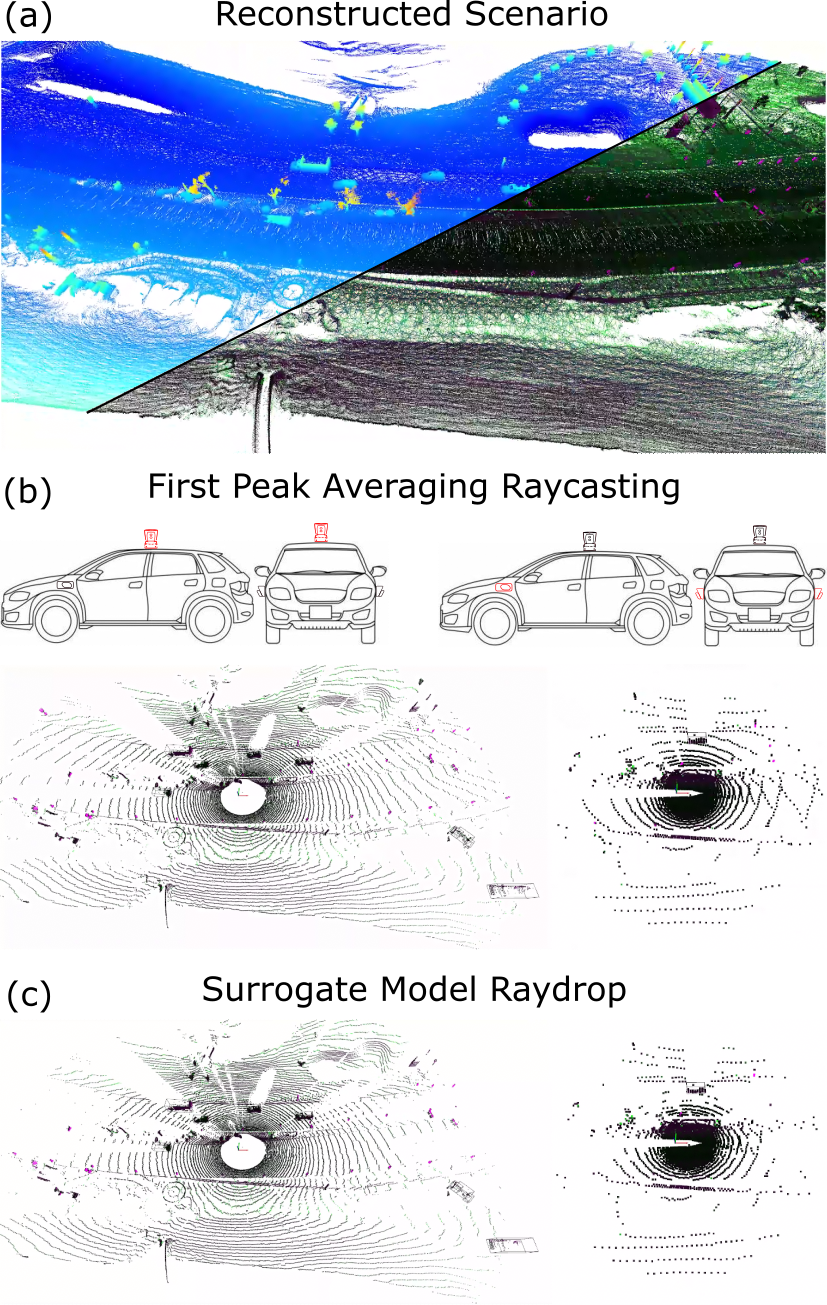}
    \caption{Overview of Simulation Pipeline}
    \label{fig:pipeline_overview}
\end{figure}

\section{Methodology}
\subsection{3D Scene Reconstruction}
\label{sec:reconstruction}
Given a sequence of single frame point clouds, ground truth bounding box annotations are used to crop out foreground object points. If an instance moves more than 0.5 meters in the global frame,  the instance is treated as a dynamic object. Since annotation for dynamic objects are less accurate, the bounding box dimensions are slightly enlarged before cropping, to ensure complete removal of all foreground points. Using odometry, single frame point cloud without foreground objects are transformed into the global coordinate frame and subsequently accumulated to obtain a dense 3D reconstruction of the sequence. Voxel downsampling and point cloud radius outlier removal are performed as post-processing steps, in order to reduce memory requirements and remove noisy points. In the case where odometry is inaccurate, \ac{SLAM} and \ac{ICP} \cite{besl1992method} can be used to improve mapping quality.

For object reconstruction, all bounding boxes belonging to the same object throughout the sequence can be identified. The cropped point cloud cluster from each frame can be returned to x-axis aligned origin using bounding box location and orientation. To reduce the impact of human annotation and odometry imperfections, a generalized \ac{ICP} can be used to improve the alignment. However, if the source and/or target point cloud contain very few points, minimizing the point cloud distance using \ac{ICP}  often leads to unreasonable reconstructions. Thus, we adaptively employ \ac{ICP} if both source and target point cloud contains greater than threshold number of points, otherwise, object clusters are simply accumulated. The foreground objects can then be inserted into the background to simulate a variety of scenarios.

\subsection{Raycasting method}
\label{sec:raycasting}
Theoretically, the reconstructed 3D scene is a 2D surface embedded in the 3D space, represented in the form of a dense point cloud. Through raycasting, intersections between the laser beams and point cloud can be computed. LiDARsim rendered dense point cloud as surfels \cite{pfister2000surfels} and used Intel Embree Engine to compute ray-disc intersections \cite{moller2005fast}. Langer et al. used \ac{CP} raycasting, which projects dense point cloud into a range image, and the closest point from each pixel is extracted to render the raycasted point cloud \cite{langer2020domain}. However, localization, calibration and sensor synchronization are subject to error, points in the reconstructed point cloud do not strictly lie on 2D surface of the scene. Both raycasting methods suffer from noisy reconstructions, leading to noisy raycasted point clouds. Moreover, \ac{CP} raycasting only works with spinning scan LiDAR. For LiDARs with irregular scan pattern, such as the DJI Livox, multiple rays might land in the same pixel, leading to redundant points.

\begin{figure}[!ht]
    \centering
    \begin{tabular}{cc}
    \includegraphics[height=3.5cm]{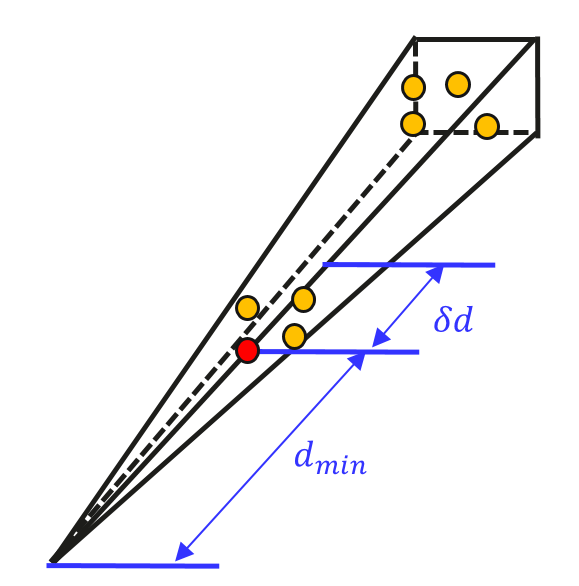} &
    \includegraphics[height=3.5cm]{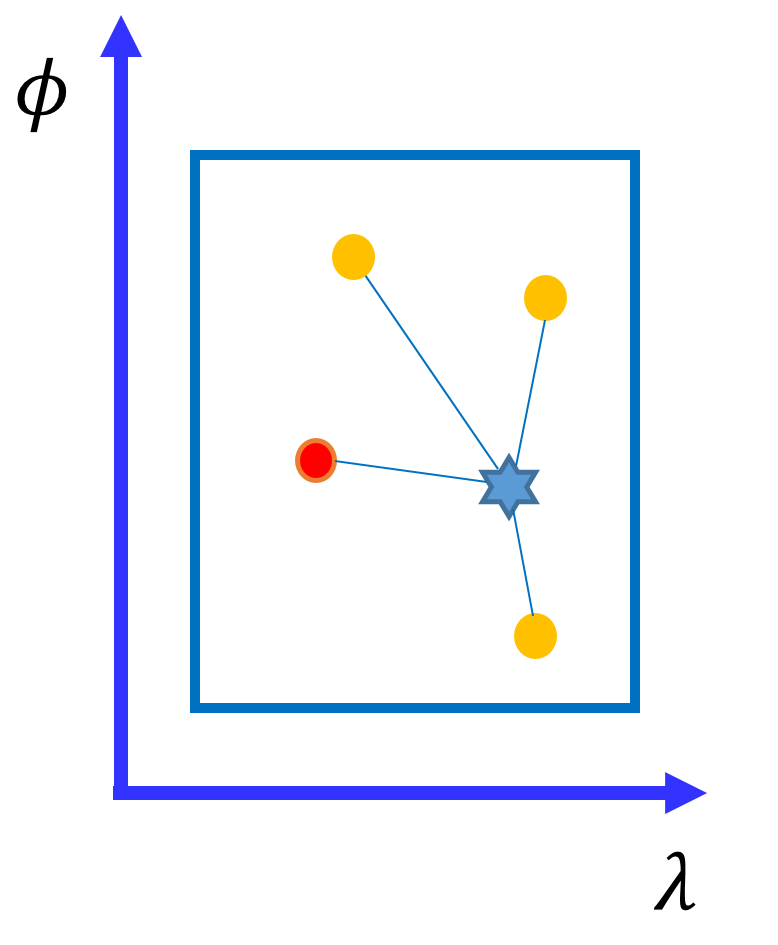} \\
    (a) & (b)
    \end{tabular}
    \caption{First peak averaging raycasting}
    \label{fig:firstpeak_idw}
\end{figure}

To solve this problem, we propose \acf{FPA} raycasting. First, the reconstructed point cloud is projected into a range image. Each pixel of the range image corresponds to a frustum in the 3D space, as shown in Figure \ref{fig:firstpeak_idw}(a). Each point within the frustum can be defined using the Spherical coordinates ($d,\lambda,\phi$) or the Cartesian coordinates ($x, y, z$). Typically, the depth distribution of all points within the frustum forms multiple peaks, due to occlusion and observation of the scene from multiple angles. The intuition behind \ac{FPA} raycasting is to average points from the closest peak, in order to estimate the true 2D surface from the noisy 3D point cloud. We are interested in the closest peak, because it corresponds to the first intersection between the ray and the 3D scene. To speed up the simulation, a fixed peak width $\delta d$ is used to screen the first peak points. In Figure \ref{fig:firstpeak_idw}(b), $\lambda_l$ and $\phi_l$ are the azimuth and elevation of the casted laser beam (indicated by the blue star),  $\lambda_i$ and $\phi_i,~i=1,2,\ldots,N$ are the azimuth and elevation of each point within the closest peak (indicated by the red and yellow circles, with the red circle being the closest point within the frustrum). The intersection between the ray and the 3D scene is estimated with the \ac{IDW} averaging. 
\begin{equation}
    f = \sum_{i=0}^N w_i f_i
\end{equation}
where $f$ can be the $x$, $y$, $z$ coordinates or the feature, such as the intensity, of a point. The inverse distance weights $w_i$ and inverse distance $d_i$ are defined as:
\begin{equation}
    w_i = \frac{\tilde{d}_i}{\sum_{j=1}^N \tilde{d}_j}
\end{equation}
\begin{equation}
    \tilde{d}_i = \frac{1}{\sqrt{(\lambda_i-\lambda_l)^2+(\phi_i-\phi_l)^2}}
\end{equation}

\subsection{Raydrop}
\label{sec:raydrop}
Laser returns are affected by many factors, such as the distance, incidence angle, material reflectivity and atmospheric composition. The \ac{FPA} raycasting algorithm records every ray-scene intersection, without taking into account the aforementioned factors. Thus, a domain gap exists, where the simulation point cloud contains more points than the real point cloud. LiDARsim developed a 2D UNet for raydrop \cite{ronneberger2015u}, trained using pairs of simulation and real point cloud range images. This requires strict pixel-wise correspondence between the two range images, and it cannot be guaranteed due to odometry and calibration error. Furthermore, deep learning models require large amounts of data to generalize. With convolution layers, the encoder-decoder network learns to drop rays based on context. Generalization to different geographies will be limited. A real LiDAR, however, does not use the wide perceptive field of the UNet to determine laser return. To address these problems, we propose a surrogate model (MLP) for the laser head.
\begin{equation}
    r = MLP(d,\theta,i) \label{eq:mlp}
\end{equation}
where $r$ is a ray's return probability, $d$, $\theta$, $i$ are the distance, incidence angle and simulation intensity of the ray. By learning ray-wise drop/return probability, the surrogate model does not require pixel-wise range image correspondence between the real and the simulation point clouds. Furthermore, the model does not require large amounts of data to train and can generalize across geographies.

\begin{figure}[!ht]
    \centering
    \begin{tabular}{cc}
    \includegraphics[height=3cm]{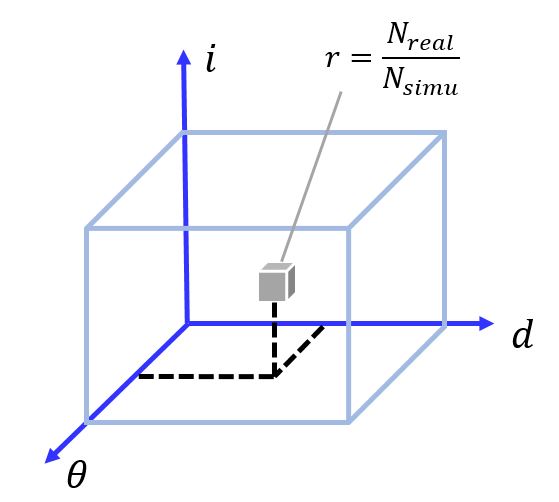} &
    \includegraphics[height=3cm]{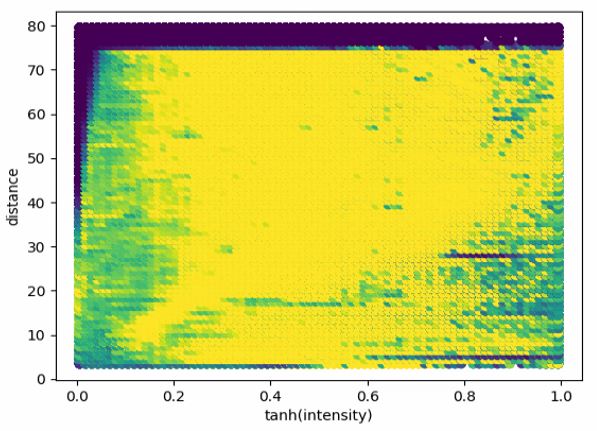} \\
    (a) & (b)
    \end{tabular}
    \caption{Input parameter space of the surrogate model}
    \label{fig:parameter_space}
\end{figure}

Given a pair of raycasted and real dataset, the $d$, $\theta$ and $i$ of each point can be computed and projected into the parameter space, as shown in Fig. \ref{fig:parameter_space}(a).
\begin{equation}
    d = \sqrt{x^2+y^2+z^2}
\end{equation}
where $x, y, z$ is Cartesian coordinate of a raycasted point in the LiDAR frame
\begin{equation}
    \theta = \arccos{\frac{\vec{R} \cdot \vec{N}}{\|\vec{R}\| \|\vec{N}\|}}
\end{equation}
where $\vec{R}$ is the ray vector, $\vec{N}$ is the point's normal vector, computed from single frame point cloud using Open3D \cite{zhou2018open3d}. Intensity, $i$, is taken directly from the real point cloud. For the \ac{FPA} simulation point cloud, the raycasted intensity is the average reflection intensity of laser beams from different directions, which abstractly describes the material's reflectivity.

To train the model, we pair the real dataset with a simulation dataset, which is raycasted using the real dataset's LiDAR poses. With sufficient data, the parameter space will be filled, which can then be voxelized. The ratio between the number of real points and the number of simulation points, $r$, can be computed for each voxel. This represents the probability the ray would be returned, at the given $d$, $\theta$, $i$. The parameter space can be converted directly into a lookup table for inference. We approximate the parameter space with an MLP, to enable GPU inference acceleration and infer values for voxels without data.

Figure \ref{fig:parameter_space}(b) shows an example 2D projection of the voxelized input parameter space. Each pixel shows the ray's return probability, with yellow and purple representing a probability of 1 and 0 respectively. A LiDAR point cloud cutoff at 75 meters is observed, and the rays that intersect farther away from the LiDAR require greater intensity to be returned. The parameter space is not very smooth, due to voxelization. In order to reduce the influence of the unsmoothness, the data of a voxel will not be used for training if the number of simulation points in the voxel is less than the given threshold or the number of the real points is larger than that of the simulation points.

\subsection{\acf{LPCS}}
Raycasting algorithm requires the selection of three hyperparameters, the range image width \& height and averaging peak width. To tune hyperparameters, a point cloud similaity metric is required. By computing the sum of squared distances between nearest neighbors, \ac{CD} does not take into account mismatching local density. Such low-level features, however, will be extracted by backbones of the perception architectures and influence model prediction. Inspired by \ac{LPIPS}, we propose \ac{LPCS}, which compares the backbone feature distance between pairs of point clouds from a model pretrained on real data. Given a pair of corresponding point cloud, all objects of interest can be cropped out using bounding box annotations, yielding two subset point clouds for simulation and real, $P_{s}, P_{r} \in \mathbb{R}^3$, respectively. \ac{LPCS} can be computed via the absolute element-wise distance between the feature of both subsets:
\begin{equation}
    LPCS(P_s, P_r) = \sum_{i=1}^{N}|F(P_{s})_i - F(P_{r})_i|
\end{equation}
where $F$, the model backbone, outputs a high-dimensional vector of $N$ terms

\subsection{Novel Sensor Synthesis}
\label{sec:sensor_aug}
With reconstructed maps and objects, a dense representation of any single frame can be obtained by inserting the objects using their poses from the real frame. By providing the raycasting algorithm with 1. the elevation and azimuth angle of each ray with respect to the LiDAR frame 2. the pose of the LiDAR with respect to the global frame,  we can simulate a corresponding point cloud, under a new LiDAR configuration. Note that if the new LiDAR's \ac{FOV} dramatically exceed that of the original dataset, part of the new dataset might be missing. This can be circumvented by collecting the original dataset using the LiDAR with largest \ac{FOV}, and optionally, highest density. Our simulation pipeline greatly reduces the data collection and annotation costs to train a model that can generalize to any combination of car and LiDAR models.

\subsection{Pedestrian Simulation}
\label{sec:VRU}
Previous works on simulation for object detection focused primarily on rigid object classes, such as car. We attempt to expand simulation classes to \ac{VRU}, in particular, pedestrians. Following the same reconstruction procedure from Section \ref{sec:reconstruction}, a pedestrian object library is generated. However, body movements lead to deformed reconstructions. To address this problem, we employ the \ac{SMPL}\cite{loper2015smpl} dataset, which provides realistic 3D human CAD models. Unlike cars, pedestrians share similar geometry and mainly differ in their postures. Thus, we believe that the \ac{SMPL} models provide sufficient degrees of freedom to capture the diverse set of pedestrian poses that can be found on the road. During raycasting, CAD models are converted to point clouds by sampling a large number of points on the surface of triangle meshes. A tight bounding box can also be generated to enclose all points. Using statistics, dimensions of the real bounding boxes and its enclosed point cloud cluster can be computed. The distributions are used to filter out unrealistic CAD poses, as well as loosening the bounding box annotations to match human annotations.

\section{Experimental Evaluation}
Through the following experiments, we intend to demonstrate the fidelity and potential applications of our pipeline.
\begin{enumerate}
    \item To verify the fidelity of our raycasting and raydrop algorithms, we show that object detection models trained using real data or simulation data, achieves similar performance when evaluated on the real validation data.
    \item To prove \ac{LPCS} as an useful point cloud similarity metric, we show negative correlation between \ac{LPCS} and object detection \ac{mAP}
    \item To demonstrate sensor type conversion, we show that with only source sensor data and 10\% annotated target sensor data, we can simulate target sensor and achieve similar object detection performance as 100\% annotated target sensor data.
    \item To demonstrate \ac{VRU} simulation, we provide a baseline for pedestrian simulation and compare CAD models against reconstructed pedestrian.
\end{enumerate}

\subsection{Experimental Setup}
\subsubsection{Dataset}
We evaluate our simulation pipeline using Waymo Open Dataset, which consists of 800 training and 200 validation segments spanning over different geographical locations \cite{sun2020scalability}. Each segment provides synchronized sensor data over 20 seconds. Each LiDAR frame is composed of 5 LiDARs: top, left, right, front and rear, captured at 10Hz. For all of our experiments, we perform reconstruction and mapping using top LiDAR, due to its high density and long range of 75 meters. 
\subsubsection{Object Detection Model}
OpenPCDet \cite{openpcdet2020} implementation of Centerpoint \cite{yin2021center} will be used to evaluate the quality of our simulation dataset. During training, no data augmentation is performed, in order to minimize stochasticity among trials. All models are trained over 80 epochs.

\subsection{First Peak Averaging Raycasting Ablation}
\label{sec:fpa_ablation}
With 2D and 3D object detection, we quantitatively benchmark \ac{CP} against \ac{FPA}. Given a real LiDAR frame, a dense representation can be obtained via reconstruction. Two simulation datasets, which are identical to the real dataset, can be raycasted using \ac{CP} and \ac{FPA}. Both datasets employ the same LiDAR pose and ray configurations from Waymo Open Dataset's calibration. Centerpoint is trained using the \ac{CP}, the \ac{FPA} and the real dataset. The trained models are evaluated on the real validation dataset, which is summarized in Table \ref{tab:raycastablation}. When trained using only XYZ, \ac{FPA} method provides a small performance gain of $+0.35/+0.05\%$ and $+0.91/+1.31\%$ for 3D and 2D $mAP_{0.5/0.7}$. This shows that the \ac{FPA}, compared to \ac{CP}, reduces noise in the simulated point clouds. When point cloud features (intensity, elongation) are included during training, \ac{FPA} provides $+1.72/+1.85\%$, $+1.10/+1.89\%$ improvement for 3D and 2D $mAP_{0.5/0.7}$. More importantly, the inclusion of point cloud features lead to improved model performance for \ac{FPA}, but worsened 3D $mAP_{0.7}$ model performance for \ac{CP}. This indicates that the simulated point cloud feature by \ac{CP} compared to \ac{FPA} is also more noisy and can cause confusion for the model.



\begin{table} \scriptsize
\centering
\caption{Top LiDAR Raycasting (Real Top Validation)} 
\subcaption*{\textbf{(a) XYZ}}
\begin{tabular}{ c c c  }
\toprule
 & 3D mAP IoU 0.5/0.7 & 2D BEV IoU 0.5/0.7 \\ 
\midrule
Simulation CP & 74.06/41.93 & 77.93/61.30\\
Simulation FPA & 74.41/41.98 & 78.84/62.61\\
Real & 77.59/47.27 & 81.19/66.01\\
\bottomrule
\end{tabular}
\subcaption*{\textbf{(b) XYZ + Intensity + Elongation}}
\vspace{6pt}
\begin{tabular}{ c c c  }
\toprule
 & 3D mAP IoU 0.5/0.7 & 2D BEV IoU 0.5/0.7 \\ 
\midrule
Simulation CP &  74.44/40.33 & 79.46/61.86  \\
Simulation FPA & 76.16/42.18 & 80.56/63.75 \\  
Real & 79.23/48.21 & 82.57/67.24 \\
\bottomrule
\end{tabular}
\label{tab:raycastablation}
\end{table}

\subsection{Raydrop Ablation}

\begin{figure}
    \centering
    \includegraphics[width=0.4\textwidth]{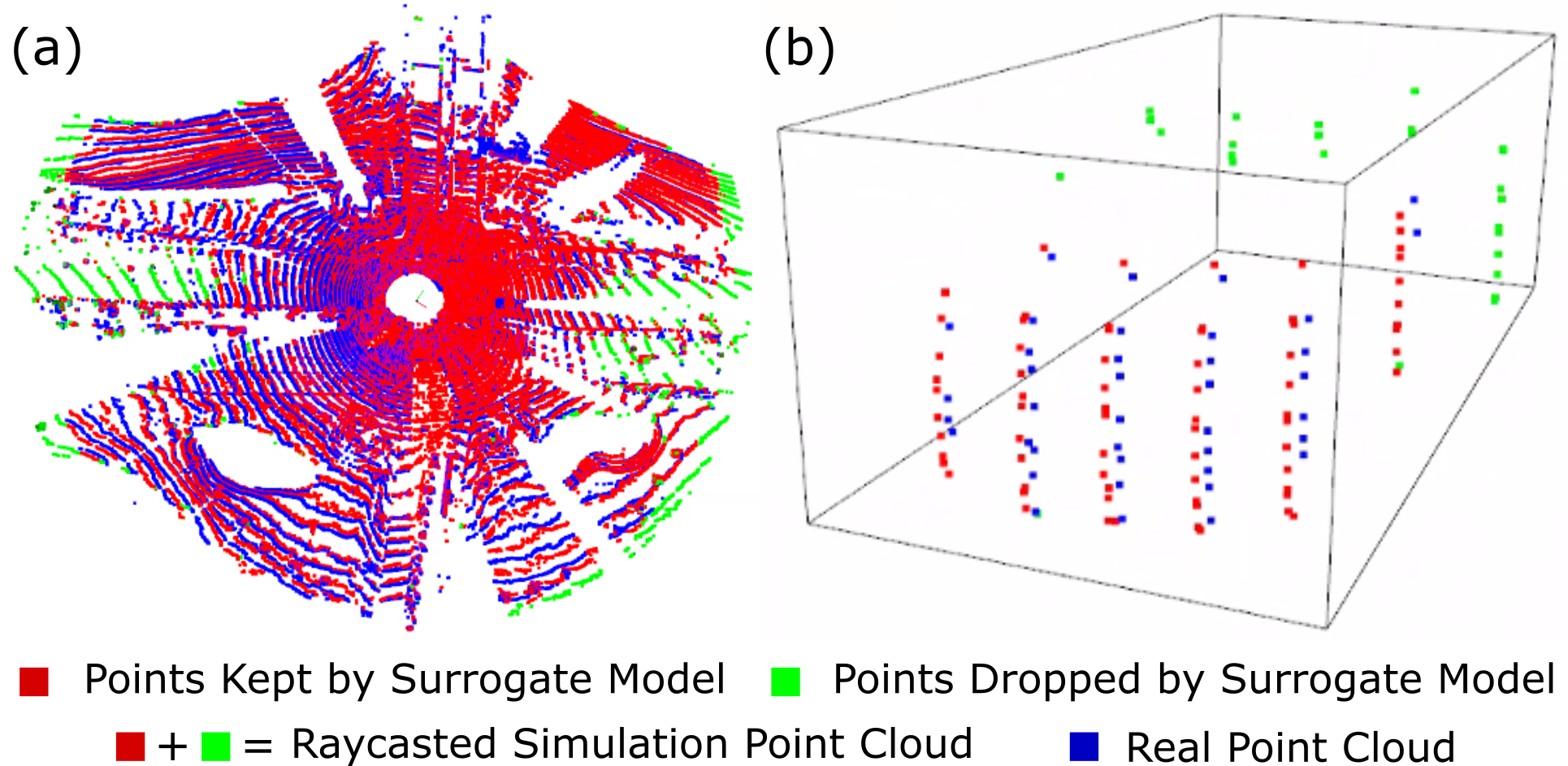}
    \caption{Surrogate Model Raydrop Visualization (a) Top LiDAR raydrop (b) Side LiDAR raydrop}
    \label{fig:raydrop_vis}
\end{figure}

To show the generalization capability of the proposed raydrop model, we will use a surrogate model trained on the validation set to drop points in the training set. Following \ref{sec:fpa_ablation}, \ac{FPA} simulation of the real validation dataset is first generated. Following Section \ref{sec:raydrop}, simulation and real validation dataset is used to train the surrogate model. For each point in the raycasted simulation training set, the surrogate model checkpoint provides a probability the point should be kept. By choosing a threshold, probabilities can be converted to binary drop/keep masks. The higher the threshold, the more points will be dropped. The trained model's performance on the real validation set is summarized in Table \ref{tab:raydropablation}. "No raydrop" indicates original raycasting results, and "Real" indicates real data. Since XYZIE is used for training, these two rows are identical to "Simulation FPA" and "Real" rows of Table \ref{tab:raycastablation}(b). At threshold $0.28$, 3D $mAP_{0.5/0.7}$ is maximized at $77.56/44.93\%$, a $+1.40/+2.75\%$ improvement compared to without raydrop. 2D $mAP_{0.5/0.7}$ also increases $+1.51/+1.49\%$ and $+0.65/+2.00\%$ at $0.30$ and $0.28$ threshold, respectively. Thus, the surrogate model is an extremely light-weight and generalizable method to effectively reduce the sim2real domain gap. Figure \ref{fig:raydrop_vis}(a) provides a visual comparison of point cloud with and without raydrop. After raydrop, simulation point cloud more closely matches that of the real point cloud.


\begin{table} \scriptsize
\begin{center}
\caption{Top LiDAR Raydrop (Real Top Validation)}
\begin{tabular}{ c c c  }
\toprule
Threshold & 3D mAP IoU 0.5/0.7 & 2D BEV IoU 0.5/0.7 \\ 
\midrule
No raydrop & 76.16/42.18 & 80.56/63.75 \\  
 0.28 & 77.56/44.93 & 81.21/65.75 \\
 0.30 & 77.36/42.80 & 82.07/65.24 \\
 0.32 & 77.22/43.63 & 81.30/64.80 \\
 0.34 & 76.63/41.24 & 80.51/63.02 \\
Real & 79.23/48.21 & 82.57/67.24 \\
\bottomrule
\end{tabular}
\label{tab:raydropablation}
\end{center}
\end{table}

\subsection{Learned Point Cloud Similarity}
We perform a small hyperparameter search using 10\% of the validation set. Using different combinations of hyperparameters (averaging peak width, range image width, range image height), new simulation validation datasets are raycasted. With a real data trained model, we can obtain the \ac{mAP} of each raycasted dataset, as well as computing the \ac{LPCS} between the raycasted and the real dataset.

\begin{figure}
    \centering
    \includegraphics[width=0.4\textwidth]{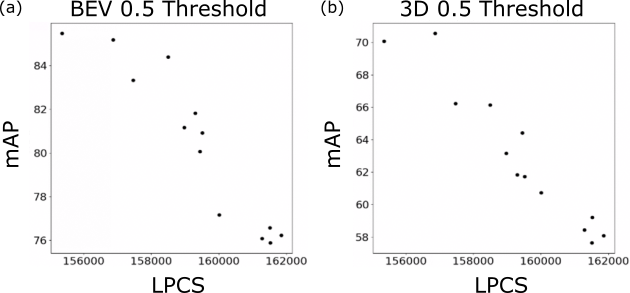}
    \caption{Correlation between LPCS and Object Detection \ac{mAP}}
    \label{fig:lpcs_res}
\end{figure}

Figure \ref{fig:lpcs_res} shows a negative correlation between \ac{LPCS} and \ac{mAP}, which implies that \ac{LPCS} provides a good estimate of the domain gap between the simulation and the real dataset. Using \ac{LPCS} to guide gridsearch of raycasting hyperparameters, we conclude that the optimal range image dimension is $2560\times128$, along with an averaging peak width of 20cm. These hyperparameters are used in all of the top LiDAR experiments.

\subsection{Novel Sensor Synthesis}
In this section, we showcase an example novel sensor synthesis, from top to side LiDARs. Following Section \ref{sec:sensor_aug}, side LiDAR dataset can be raycasted using top LiDAR reconstructions. First 10\% of the real and raycasted side LiDAR training set is used to train a surrogate raydrop model, to raydrop the entire raycasted training set. Using the first 10\% and random 10\% of the real side LiDAR training set, the simulation data trained models are further finetuned. Object detection performance, evaluated on the real side LiDAR validation set, is summarized under Table \ref{tab:sensoraug}(a). Some real data trained baselines are also provided in Table \ref{tab:sensoraug}(b). If a real top LiDAR trained model (Top 100\%) is evaluated on the real side validation set, we observe around 20\% domain gap compared to the real side LiDAR trained model (Side 100\%). 
A similar performance is observed for raycasted simulation dataset (Raycast only). However, raydrop (Raycast + Raydrop) is extremely useful. Compared to (Raycast only), we observe $+15.02/+7.80\%$ and $+13.01/+12.19\%$ improvement. Compared to (Side First 10\%), we observe 
$+1.78/+1.01\%$ and $+2.36/+6.24\%$ improvement.
Figure \ref{fig:raydrop_vis}b shows that after dropping the scanlines around the top and rear sections of the car, the raydropped point cloud is a lot more similar compared to the real point cloud than the raycasted point cloud. The reduced domain gap is likely the main contributing factor to the improved performance. 
Finetuning with first 10\% further boosts model performance (Raycast + Raydrop + First 10\%). Compared to (Side First 10\%), we observe $+6.87/+10.8\%$ and $+6.16/+9.51\%$ improvement. Compared to (Side All 100\%), we observe a  gap of $-3.19/-6.98\%$ and $-1.17/-1.93\%$. This suggests that given a large scale primary dataset of an old sensor and a small scale secondary dataset of a new sensor, our simulation pipeline can leverage both datasets and train a model that is capable of achieving similar performance compared to a model trained on the large scale dataset of the new sensor. Alternatively, the simulation pipeline can be thought of as a data augmentation generator. The large scale primary dataset can be converted to augmentation frames for the secondary dataset, to improve the model's generalization capability on the secondary dataset.

\begin{table} \scriptsize
\begin{center}
\caption{Side LiDAR Simulation (Real Side Validation)}
\subcaption*{\textbf{(a) Simulation Ablation}}
\begin{tabular}{ c c c c c c  }
\toprule
Raycast & Raydrop & Finetune & \begin{tabular}[c]{@{}l@{}}3D mAP\\ IoU 0.5/0.7\end{tabular}  & \begin{tabular}[c]{@{}l@{}}2D BEV\\ IoU 0.5/0.7\end{tabular} \\ 
\midrule
\cmark & \xmark & \xmark & 61.98/28.25 & 68.98/51.00 \\
\cmark & \cmark & \xmark & 77.00/36.05 & 81.99/63.19 \\
\cmark & \cmark & First 10\% & 82.09/45.86 & 85.79/66.46\\
\cmark & \cmark & Random 10\% & 82.99/49.90 & 85.77/69.45\\
\bottomrule
\end{tabular}
\subcaption*{\textbf{(b) Real Data Baselines}}
\vspace{6pt}
\begin{tabular}{ c c c c }
\toprule
Dataset Sensor & Volume & \begin{tabular}[c]{@{}l@{}}3D mAP\\ IoU 0.5/0.7\end{tabular}  & \begin{tabular}[c]{@{}l@{}}2D BEV\\ IoU 0.5/0.7\end{tabular} \\ 
\midrule
Top & All 100\% & 62.31/32.39 & 65.02/51.70\\
Side & First 10\% & 75.22/35.06 & 79.63/56.95\\
Side & Random 10\% & 83.77/49.57 & 86.96/68.39\\
Side & All 100\% & 85.28/52.84 & 88.02/70.78\\
\bottomrule
\end{tabular}
\label{tab:sensoraug}
\end{center}
\end{table}

\subsection{Deformable Object Simulation}
\begin{figure}
    \centering
    \includegraphics[width=0.45\textwidth]{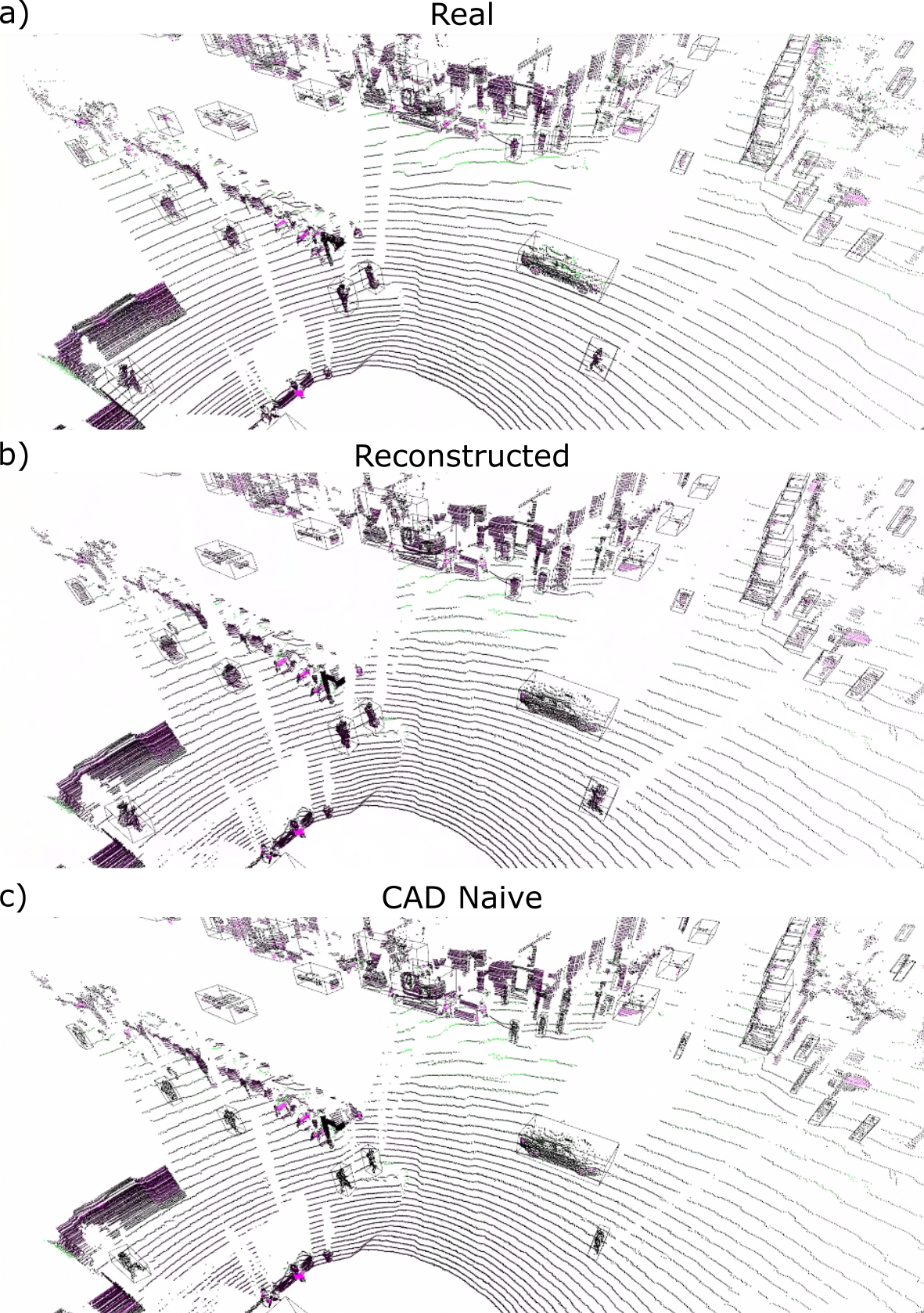}
    \caption{Pedestrian Simulation Visualization}
    \label{fig:ped_vis}
\end{figure}

Table \ref{tab:pedablation} compares the pedestrian simulation using reconstructed and CAD pedestrians, when evaluated on the real top validation set. "Real" represents real data trained model. "CAD Naive" represents randomly sampling from the CAD library and replacing real pedestrians. "CAD Modified" improves upon "CAD Naive" with the inclusion of pose filtering and bounding box adjustment as described in Section \ref{sec:VRU}. "Reconstruction" replaces real with reconstructed pedestrians. Figure \ref{fig:ped_vis} provides a visual comparison of the above experiments. The reconstructed pedestrians leads to noticeably thicker outline compared to the real and CAD point clouds.

Compared to car classes, reconstructed pedestrians suffer from increased sim-real domain gap of $-8.55/-14.48\%$ and $-8.75/-12.87\%$ for 3D and 2D $mAP_{0.5/0.7}$. "CAD Modified" achieves the lowest domain gap of $-3.75/-7.31\%$ and $-4.05/-5.84\%$. Our results provides a baseline for \ac{VRU} simulation and show that \ac{SMPL} is a viable alternative to replace reconstruction for pedestrians.

\begin{table} \scriptsize
\begin{center}
\caption{Pedestrian Simulation (Real Top Validation)}
\begin{tabular}{ c c c  }
\toprule
Training Dataset & 3D mAP IoU 0.25/0.50 & 2D BEV IoU 0.25/0.50 \\ 
\midrule
CAD Naive & 55.62/06.56 & 59.31/11.92 \\
Reconstruction & 69.76/47.14 & 70.49/55.44 \\
CAD Modified & 74.56/54.31 & 75.19/62.47 \\
Real & 78.31/61.62 & 79.24/68.31 \\
\bottomrule
\end{tabular}
\label{tab:pedablation}
\end{center}
\end{table}

\section{Conclusion}
In this work, we propose a point cloud based simulation pipeline. Experiments show that the pipeline is capable of transferring a dataset collected by an old sensor to recreate the LiDAR stream that would have been collected by a new sensor configuration of different density, placement and scanning mechanism. It greatly reduces the cost of data collection and annotation to generalize model performance from a particular vehicle-LiDAR setup to any desired combination. To harness the full potential of our pipeline, we look to close the domain gap between simulated and real pedestrians by augmenting the \ac{SMPL} dataset with accessories, such as backpacks and handbags, as well as expanding \ac{VRU} simulation to cyclists.

{\small
\bibliographystyle{IEEEtran}
\bibliography{summarybib}
}

\end{document}